\documentclass[sigconf]{acmart}

\setcopyright{acmlicensed}
\copyrightyear{2024}
\acmYear{2024}
\acmConference[CIKM '24] {Proceedings of the 33rd ACM International Conference on Information and Knowledge Management}{October 21--25, 2024}{Boise, ID, USA.}
\acmBooktitle{Proceedings of the 33rd ACM International Conference on Information and Knowledge Management (CIKM '24), October 21--25, 2024, Boise, ID, USA}
\acmISBN{979-8-4007-0436-9/24/10}
\acmDOI{10.1145/3627673.3679532}

\usepackage{subfigure}
\usepackage{booktabs}
\usepackage{multirow}
\usepackage{graphicx}
\usepackage{amsmath,amsfonts,bm}
\usepackage{float}
\usepackage{enumitem}
\usepackage[normalem]{ulem}
\useunder{\uline}{\ul}{}
\usepackage{bbm}
\usepackage[linesnumbered,ruled,vlined]{algorithm2e}
\newtheorem{definition}{Definition}
\newtheorem{problem}{Problem}

\begin{document}
\begin{sloppypar}
\title{MTSCI: A Conditional Diffusion Model for Multivariate Time Series Consistent Imputation}


\author{Jianping Zhou}
\affiliation{%
  \institution{Shanghai Jiao Tong University}
  \city{Shanghai}
  \country{China}
}
\email{jianpingzhou@sjtu.edu.cn}

\author{Junhao Li}
\affiliation{%
  \institution{Shanghai Jiao Tong University}
  \city{Shanghai}
  \country{China}
  }
\email{Lijunhao\_hz@sjtu.edu.cn}

\author{Guanjie Zheng}
\authornote{Corresponding author}
\affiliation{%
  \institution{Shanghai Jiao Tong University}
  \city{Shanghai}
  \country{China}}
\email{gjzheng@sjtu.edu.cn}

\author{Xinbing Wang}
\affiliation{%
 \institution{Shanghai Jiao Tong University}
 \city{Shanghai}
 \country{China}}
\email{xwang8@sjtu.edu.cn}

\author{Chenghu Zhou}
\affiliation{%
  \institution{Chinese Academy of Sciences}
  \city{Beijing}
  \country{China}}
\email{zhouch@Ireis.ac.cn}


\begin{abstract}
Missing values are prevalent in multivariate time series, compromising the integrity of analyses and degrading the performance of downstream tasks.
Consequently, research has focused on multivariate time series imputation, aiming to accurately impute the missing values based on available observations.
A key research question is how to ensure imputation consistency, i.e., \textit{intra-consistency} between observed and imputed values, and \textit{inter-consistency} between adjacent windows after imputation. 
However, previous methods rely solely on the inductive bias of the imputation targets to guide the learning process, ignoring imputation consistency and ultimately resulting in poor performance.
Diffusion models, known for their powerful generative abilities, prefer to generate consistent results based on available observations.
Therefore, we propose a conditional diffusion model for \underline{\textbf{M}}ultivariate \underline{\textbf{T}}ime \underline{\textbf{S}}eries \underline{\textbf{C}}onsistent \underline{\textbf{I}}mputation (MTSCI).
Specifically, MTSCI employs a contrastive complementary mask to generate dual views during the forward noising process. 
Then, the intra contrastive loss is calculated to ensure intra-consistency between the imputed and observed values.
Meanwhile, MTSCI utilizes a mixup mechanism to incorporate conditional information from adjacent windows during the denoising process, facilitating the inter-consistency between imputed samples.
Extensive experiments on multiple real-world datasets demonstrate that our method achieves the state-of-the-art performance on multivariate time series imputation task under different missing scenarios.
Code is available at \url{https://github.com/JeremyChou28/MTSCI}.
\end{abstract}



\maketitle

\section{Introduction}
Multivariate time series data widely exists in various real-world applications, e.g., transportation~\cite{ming2022traffic,tedjopurnomo2020traffic_survey}, meteorology~\cite{han2021climate,abhishek2012weather}, healthcare~\cite{tonekaboni2020TNC}, energy~\cite{akay2007electricity}, etc.
The integrity of time series plays a crucial role on tasks such as forecasting~\cite{zeng2023dlinear,liu2023itransformer} and classification~\cite{li2022classification}. 
However, missing data is a common issue in real-world datasets due to device failures, communication interruptions, and human errors~\cite{tan2018incomplete,greco2012incomplete}, which impairs the downstream task performance and renders the integrity analysis approaches inapplicable.
Consequently, multivariate time series imputation, which aims to accurately impute missing values using available observations, arises as an important research question.

\begin{figure}[!t]
    \centering
    \includegraphics[width=\linewidth]{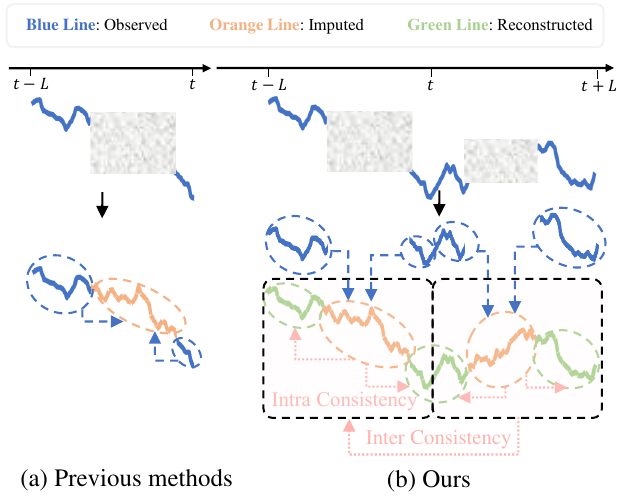}
    \caption{(a) Previous methods only use the observed values in a single window to impute missing values. (b) Our method uses the adjacent window to assist in imputation to maintain \textit{inter-consistency}, and constraints the imputed targets reciprocally reconstruct the observed values for \textit{intra-consistency}.}
    \vspace{-2mm}
    \label{fig:introduction}
\end{figure}

Early studies are based on statistical learning~\cite{fung2006mean,batista2002knn} and machine learning methods~\cite{white2011mice,lee2000alnmf,yu2016trmf}. 
Subsequently, numerous deep learning-based methods have been proposed. 
Some approaches~\cite{che2018grud,cao2018brits,shukla2020mtan,du2023saits} treat the imputation task as a deterministic point estimation problem, while others~\cite{fortuin2020gpvae,yoon2018gain,chen2023csbi,tashiro2021csdi,alcaraz2022sssd} view it as a probabilistic generative problem.
However, these methods rely solely on the inductive bias of artificially simulated imputation targets to guide the learning process, as illustrated in Figure~\ref{fig:introduction} (a), neglecting the crucial aspect of imputation consistency.

The imputation consistency can be divided into two categories: (i)\textit{\textbf{intra-consistency}} and (ii)\textit{\textbf{inter-consistency}}.
As shown in Figure~\ref{fig:introduction} (b), consider two samples of adjacent windows in incomplete time series.
\textit{Intra-consistency} implies that the imputed values, guided by observed values, should facilitate the reconstruction of observed values, ensuring consistency between imputed and observed values, thereby reducing imputation bias.
\textit{Inter-consistency} means that when imputing a single sample, the sample from the adjacent window should be considered to ensure that the \textit{complete} sample maintains temporal consistency with the adjacent window.
This concept coincides with the idea of good continuity in the temporal samples of adjacent windows~\cite{tonekaboni2020TNC}.
However, no existing imputation method addresses the issue of imputation consistency in multivariate time series imputation task. 

Nowadays, diffusion models exhibit powerful generative abilities in synthesizing images~\cite{rombach2022image}, audio~\cite{chen2020wavegradaudio}, text~\cite{gong2022diffuseqtext}, and time series~\cite{tashiro2021csdi}.
Compared to other models, diffusion models are more inclined to generate data consistent with the distribution of observed data, which aligns well with the concept of imputation consistency~\cite{ho2020ddpm}.
Therefore, to address the aforementioned issues, we propose \textbf{MTSCI}, a conditional diffusion model for \underline{\textbf{M}}ultivariate \underline{\textbf{T}}ime \underline{\textbf{S}}eries \underline{\textbf{C}}onsistent \underline{\textbf{I}}mputation.
MTSCI employs a contrastive complementary mask strategy during the noising process and a mixup mechanism that incorporates conditional information from adjacent windows, effectively guiding the model to maintain consistency and improving imputation performance.
Specifically, (i) the complementary mask strategy, which self-supervises the generation of complementary pairs of samples when simulating the imputation targets, imposes contrastive loss to teach the model to maintain \textit{intra-consistency} between the observed and imputed values.
(ii) The mixup mechanism incorporates the temporal characteristics of the adjacent window as conditional information to assist in the imputation of current window during the training stage, thus preserving \textit{inter-consistency} between adjacent windows.
Experimental results on three real-world datasets demonstrate that our method achieves state-of-the-art performance with average improvements of 17.88\% in MAE, 15.09\% in RMSE, and 13.64\% in MAPE compared to baseline methods.

In summary, the main contributions of our work are as follows:
\begin{itemize}[leftmargin=*]
    \item Motivated by the neglect of imputation consistency in existing time series imputation methods, we systematically summarize the concept of imputation consistency in multivariate time series imputation task as \textit{intra-consistency} and \textit{inter-consistency}.
    \item We propose MTSCI, a conditional diffusion model for multivariate time series consistent imputation, incorporating a complementary mask strategy and a mixup mechanism to realize intra-consistency and inter-consistency.
    \item We conduct extensive experiments on multiple real-world datasets to demonstrate the effectiveness of MTSCI. The results indicate that our method achieves the state-of-the-art performance on multivariate time series imputation task under different missing data scenarios.
\end{itemize}

\section{Related Work}
Existing multivariate time series imputation methods can be divided into four categories: \textit{statistical-based methods}, \textit{machine learning-based methods}, \textit{deterministic deep models} and \textit{probabilistic generative models}.
(i) \textit{Statistical methods}, such as Mean, Median~\cite{fung2006mean} and KNN~\cite{batista2002knn}, utilize the statistical indicators to impute missing values.
(ii) \textit{Machine learning methods} like linear imputation, state-space models~\cite{durbin2012statespace} and MICE~\cite{white2011mice} impute missing values based on linear dynamics assumptions.
Other machine learning methods like low-rank matrix factorization, e.g., NMF~\cite{lee2000alnmf}, TRMF~\cite{yu2016trmf}, TIDER~\cite{liu2022tider}, factorize incomplete data into low-rank matrices and impute missing values using the product of these matrices.
However, these methods struggle with capturing the nonlinear dynamics and handling large datasets.
(iii) \textit{Deterministic deep models} treat imputation as a deterministic point estimation problem.
For instance,
GRUD~\cite{che2018grud} uses the last observation and mean of observations to represent missing patterns.
BRITS~\cite{cao2018brits} employs bidirectional recurrent neural networks to capture temporal features.
mTAN~\cite{shukla2020mtan} and SAITS~\cite{du2023saits} design attention mechanisms to capture temporal dependencies.
TimesNet~\cite{wu2023timesnet} transforms 1D time series to 2D space to capture complex temporal variations, achieving state-of-the-art performance in multiple time series tasks, including imputation.
However, deterministic methods fall short of modeling imputation uncertainties.
(iv) \textit{Probabilistic generative models} view imputation as a missing values generation problem.
For example,
GP-VAE~\cite{kingma2013vae} combines the VAE~\cite{kingma2013vae} and Gaussian process to model incomplete time series.
GAIN~\cite{yoon2018gain} uses a GAN~\cite{goodfellow2020gan} with a hint mechanism to aid imputation.
McFlow~\cite{richardson2020mcflow} leverages normalizing flow generative models and Monte Carlo sampling for imputation.
CSDI~\cite{tashiro2021csdi} and MIDM~\cite{wang2023midm} utilize conditional diffusion models to generate missing values by treating the observed values as conditional information.
CSBI~\cite{chen2023csbi} employs the Schr\"odinger bridge algorithm for imputation.
PriSTI~\cite{liu2023pristi} extracts conditional information for spatio-temporal imputation using geographic data.
However, these methods only use the inductive bias on imputation targets to guide the learning process, which is insufficient to maintain \textit{intra-consistency} between observed and imputed values within a sequence, as well as \textit{inter-consistency} between imputed sequences.

Although there are no imputation methods considering the imputation consistency, other studies have explored consistency strategies in time series analysis.
Time series representation methods~\cite{tonekaboni2020TNC,yeche2021neighborhood} propose a consistency strategy in the sampling process, where time segments within adjacent windows exhibit high pattern consistency, such as consistent trends, periods and amplitudes.
TS2Vec~\cite{yue2022ts2vec} and TS-TCC~\cite{emadeldeen2022ts-tcc} utilize data augmentation to generate multiple views for contrastive learning, thereby learning consistent representations.
TF-C~\cite{zhang2022tfc}, CoST~\cite{woo2021cost} consider that  sequence dependencies of a time series remain consistent during transformations between the time and frequency domains.
However, these methods concentrate on complete time series, falling short in addressing multivariate time series imputation task.

\section{Preliminaries}
In this section, we define the problem of multivariate time series imputation, and then introduce the background of diffusion models and conditional diffusion models.

\subsection{Problem Definition}
\begin{definition}
    [\textbf{Multivariate Time Series.}]
    The multivariate time series denoted as $\mathbf{X} \in \mathbb{R}^{T \times C}$ contains $C$ features with length $T$.
    The mask matrix denoted as $\mathbf{M}\in\{0,1\}^{T\times C}$ indicates whether the values is observed or missing, where $\mathbf{M}^{i,j}=1$ indicates $\mathbf{X}^{i,j}$ is observed, and $\mathbf{M}^{i,j}=0$ indicates $\mathbf{X}^{i,j}$ is missing.
    Then, the observed values in $\mathbf{X}$ is denoted as $\mathbf{X}^o=\mathbf{X}\odot\mathbf{M}$, the missing values in $\mathbf{X}$ is denoted as $\mathbf{X}^m=\mathbf{X}\odot(\mathbf{1}-\mathbf{M})$.
\end{definition}

\begin{problem}
    [\textbf{Multivariate Time Series Imputation.}]
    Given the multivariate time series $\mathbf{X}$ and the mask matrix $\mathbf{M}\in\{0,1\}^{T\times C}$ over $T$ time slices, our task of multivariate time series imputation is to estimate the missing values or corresponding distributions in $\mathbf{X}$. 
    The problem can be formulated as learning a probabilistic imputation function $p_{\theta}$:
\begin{equation}
    \mathop{max}\limits_{\theta}p_{\theta}(\mathbf{X}^m|\mathbf{X}^o),\\
\end{equation}
where the goal is to approximate the real the conditional distribution $p_\theta(\mathbf{X}^m|\mathbf{X}^o)$ or minimize the estimation error on missing positions.
\end{problem}

\subsection{Diffusion Models}
A well-known diffusion model is the denoising diffusion probabilistic model (DDPM) ~\cite{ho2020ddpm}, which contains the forward noising process and the backward denoising process.
During the forward noising process, an input $\mathbf{x}_0$ is gradually corrupted to a Gaussian noise vector, which can be defined by the following Markov chain:
\begin{equation}
q(\mathbf{x}_T|\mathbf{x}_0):=\prod_{k=1}^{T}q(\mathbf{x}_k|\mathbf{x}_{k-1}), q(\mathbf{x}_k|\mathbf{x}_{k-1}):=\mathcal{N}(\sqrt{1-\beta_k}\mathbf{x}_{k-1},\beta_k\mathbf{I}),
\end{equation}
where $\beta_k\in[0,1]$ represents the noise level.
Then, sampling of $\mathbf{x}_k$ can be written as $q(\mathbf{x}_k|\mathbf{x}_0)=\mathcal{N}(\mathbf{x}_k;\sqrt{\Bar{\alpha}_k}\mathbf{x}_0,(1-\Bar{\alpha}\mathbf{I})$, where $\Bar{\alpha}_k=\prod_{s=1}^k\alpha_s$, and $\alpha_k=1-\beta_k$.
Thus, $\mathbf{x}_k$ can be simply obtained as:
\begin{equation}
\label{forward_noising}
\mathbf{x}_k=\sqrt{\Bar{\alpha}_k}\mathbf{x}_0+\sqrt{1-\Bar{\alpha}_k}\epsilon,
\end{equation}
where $\epsilon$ is a Gaussian noise. 
During the backward denoising process, DDPM considers the following specific parameterization of $p_\theta(\mathbf{x}_{k-1}|\mathbf{x}_k)$:
\begin{equation}
p_\theta(\mathbf{x}_{k-1}|\mathbf{x}_k)=\mathcal{N}(\mathbf{x}_{k-1};\mu_\theta(\mathbf{x}_k,k),\sigma_\theta(\mathbf{x}_k,k)),
\end{equation}
where the variance $\sigma_\theta(\mathbf{x}_k,k)$ is usually fixed as $\sigma^2\mathbf{I}$ and the mean $\mu_\theta(\mathbf{x}_k,k)$ is defined by a denoising network ($\mathbf{x}_\theta$ or $\mathbf{\epsilon}_\theta$).
For noising estimation, the denoising network $\epsilon_\theta$ predicts the noise, and then obtains the mean $\mu_\theta(\mathbf{x}_k,k)$:
\begin{equation}
\mu_\theta(\mathbf{x}_k,k)=\frac{1}{\sqrt{\alpha_k}}\mathbf{x}_k-\frac{1-\alpha_k}{\sqrt{1-\Bar{\alpha}_k}\sqrt{\alpha}_k}\epsilon_\theta(\mathbf{x}_k,k).
\end{equation}
The denoising network $\epsilon_\theta$ is trained by minimizing the loss $\mathcal{L}_\epsilon$:
\begin{equation}
\mathcal{L}_\epsilon=\mathbb{E}_{k,\mathbf{x}_0,\epsilon}\Vert\epsilon-\epsilon_\theta(\mathbf{x}_k,k)\Vert_2^2.
\end{equation}
For $\mathbf{x}_0$ estimation, the denoising network $\mathbf{x}_\theta$ predicts the value $\mathbf{x}_0$, and then obtains the mean $\mu_\theta(\mathbf{x}_k,k)$:
\begin{equation}
\mu_\theta(\mathbf{x}_k,k)=\frac{\sqrt{\alpha}_k(1-\Bar{\alpha}_{k-1})}{1-\Bar{\alpha}_k}\mathbf{x}_k+\frac{\sqrt{\Bar{\alpha}_{k-1}}\beta_k}{1-\Bar{\alpha}_k}\mathbf{x}_\theta(\mathbf{x}_k,k).
\end{equation}
The denoising network $\mathbf{x}_\theta$ is trained by minimizing the loss $\mathcal{L}_\mathbf{x}$:
\begin{equation}
\mathcal{L}_\mathbf{x}=\mathbb{E}_{k,\mathbf{x}_0,\epsilon}\Vert \mathbf{x}_0-\mathbf{x}_\theta(\mathbf{x}_k,k)\Vert_2^2.
\end{equation}

\subsection{Conditional Diffusion Models}
The multivariate time series imputation task aims to impute missing values $\mathbf{X}^m\in\mathbb{R}^{T\times C}$ based on the conditional information of observed values $\mathbf{X}^o\in\mathbb{R}^{T\times C}$.
The forward noising process and backward denoising process of conditional diffusion model to impute missing values are defined as follows:
\begin{equation}
p_\theta(\mathbf{x}^m_0|\mathbf{x}^o_0):=p(\mathbf{x}^m_T)\prod_{k=1}^Tp_\theta(\mathbf{x}^m_{k-1}|\mathbf{x}^m_k,\mathbf{c}), \mathbf{x}^m_T~\in\mathcal{N}(0,\mathbf{I}),
\end{equation}
\begin{equation}
\label{denoising_imputation}
p_\theta(\mathbf{x}^m_{k-1}|\mathbf{x}^m_{k},\mathbf{x}^o_0):=\mathcal{N}(\mathbf{x}^m_{k-1};\mu_\theta(\mathbf{x}^m_k,k|\mathbf{c}),\sigma_\theta(\mathbf{x}^m_k,k|\mathbf{c})\mathbf{I}),
\end{equation}
where 
$\mathbf{c}=\mathcal{F}(\mathbf{x}^o_0)$ is the conditional information output of the conditioning network $\mathcal{F}$. 
By repeatedly running the denoising step in (\ref{denoising_imputation}) till $k=1$, the imputed value $\hat{\mathbf{x}}^m_0$ is obtained.
\section{Methodology}
In this section, we elaborate on our model, MTSCI, which is designed to tackle the multivariate time series imputation task.
We start with an overview of MTSCI in Section~\ref{Overview}. 
Then, we introduce the contrastive consistency on forward noising process in Section~\ref{complementary_mask}.
Following that, we explain the consistency-assured denoising process in Section~\ref{denoising_process}.
Finally, we outline the algorithm procedures for both the training and inference stages.

\begin{figure*}[htbp]
    \centering
    \includegraphics[width=0.9\linewidth]{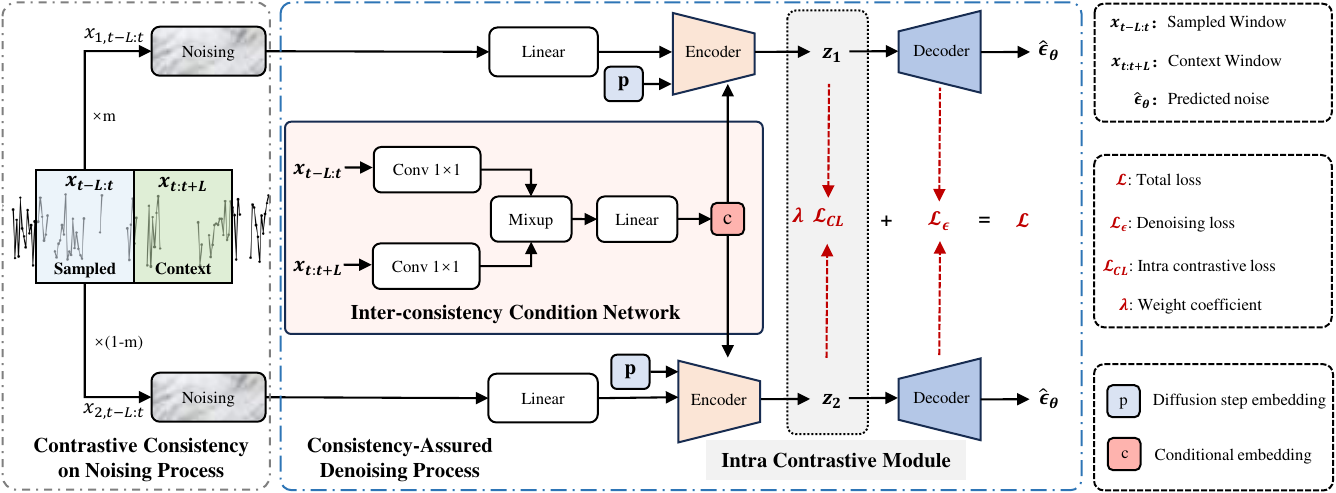}
    \caption{Overview of MTSCI. The intra contrastive module is used to maintain intra-consistency between imputed and observed values. The inter-consistency condition network is utilized to facilitate the inter-consistency between adjacent windows. The model is optimized using a combination of denoising loss $\mathcal{L}_\theta$ and intra contrastive loss $\mathcal{L}_{CL}$.}
    \label{fig:methodology}
\end{figure*}

\subsection{Overview}
\label{Overview}
MTSCI is a conditional diffusion model for multivariate time series consistent imputation.
As shown in Figure~\ref{fig:methodology}, the "sampled" window $x_{t-L:t}$ is sampled for imputation. 
First, we use contrastive complementary mask to generate two views for contrastive consistency on noising process.
Then, we utilize the intra contrastive module and inter-consistency condition network for consistency-assured denoising process, where the intra contrastive module is to calculate contrastive loss between two views, ensuring the intra-consistency between imputed and observed values.
The inter-consistency condition network is to incorporate the conditional information from "context" window for consistency between adjacent windows.

\subsection{Contrastive Consistency on Noising Process}
\label{complementary_mask}
\noindent{\textbf{Goal:}} Previous works directly use a random mask strategy to generate imputation targets and utilize the remaining observations to impute them.
However, this approach does not guarantee consistency between the imputed and the observed values, meaning the imputed values cannot guide the network to reconstruct the observed values.
This results in overfitting some outliers and then leads to deviations between the imputed values and their ground-truth.
To address this, we utilize a contrastive complementary mask strategy for the "sampled" window of the incomplete time series, generating a pair of samples where the imputation targets and observations are complementary.
This approach ensures that MTSCI learns to impute missing values consistent with the observed values during training.

\subsubsection{Contrastive Complementary Mask}
We refer to the input into the MTSCI as the "sampled" window, denoted as $x_{t-L:t}$, and the "context" window is denoted as $x_{t:t+L}$.
Using the complementary mask strategy, we generate two views of the "sampled" window. 
Based on a random mask matrix $m$ from self-supervised mask modeling process, we consider $x_{1,t-L:t}=m\odot x_{t-L:t}$ and $x_{2,t-L:t}=(1-m)\odot x_{t-L:t}$ as the two views, where $\odot$ denotes element-wise production. 
The selection of mask patterns and mask ratios in the self-supervised mask modeling process is discussed in Section~\ref{experimental_setup}.

\subsubsection{Noising}
\label{noising}
Take the $x_{1,t-L:t}$ as an example, the imputation target and conditional observation are denoted as $\mathbf{x}_0^{ta}=x_{1,t-L:t}^{ta}$, $\mathbf{x}_0^{co}=x_{1,t-L:t}^{co}$, respectively.
Based on (\ref{forward_noising}), we obtain the noised $\mathbf{x}_k^{ta
}$:
\begin{equation}
\label{noising_target}
\mathbf{x}_k^{ta}=\sqrt{\Bar{\alpha}_k}\mathbf{x}_0^{ta}+\sqrt{1-\Bar{\alpha}_k}\epsilon,
\end{equation}
where $\epsilon$ is sampled from $\mathcal{N}(0,\mathbf{I})$ with the same size as $\mathbf{x}_k^{ta
}$.

\subsection{Consistency-Assured Denoising Process}
\label{denoising_process}
\noindent{\textbf{Goal:}}
The goal is to introduce how consistency strategies can be ensured in the backward denoising process to enhance imputation performance.
We first introduce the intra contrastive loss to constrain the two views generated in Section ~\ref{complementary_mask}, enabling the imputed and observed values to reconstruct each other(\textit{\textbf{intra consistency}}). 
Subsequently, we utilize the "context" window to provide supplemental conditional information, combined with the observed conditional information in the "sampled" window via a mixup mechanism, to teach the model to impute the "sampled" window while considering the "context" window(\textit{\textbf{inter-consistency}}).
\subsubsection{Intra Contrastive Module(\textbf{Intra-consistency})}
The noised contrastive pairs obtained from Section~\ref{noising} are input into the denoising network, producing the embedding $z_1$and $z_2$ after the encoder.
Then, we calculate the representation similarity between $z_1$ and $z_2$.
To be specific, suppose that the number of samples in one batch is $N$, then after applying the complementary mask, we have $2N$ views of samples.
Inspired by the cross-entropy formulation of contrastive loss in previous works~\cite{emadeldeen2022ts-tcc,lee2023pits}, we calculate the intra contrastive loss as follows:
\begin{equation}
\label{L_intra}
    \mathcal{L}_{CL}=\frac{1}{N}\log\frac{exp(sim(z_1^i,z_2^i)/\tau)}{\sum_{m=1}^{2N}\mathbbm{1}_{[m\neq i]}exp(sim(z_1^i,z_2^m)/\tau},
\end{equation}
where $sim(u,v)=u^Tv/\Vert u\Vert \Vert v\Vert$ is the cosine similarity, $\mathbbm{1}_{m\neq i}\in \{0,1\}$ is an indicator function, $\tau$ is a temperature parameter.
By constraining the representation similarity between two views generated by complementary mask strategy, the model learns to ensure that the imputed values can also reconstruct the observed values during the imputation process.

\subsubsection{Inter-consistency Condition Network(\textbf{Inter-consistency})}
Existing methods only use the observed values to impute imputation targets in "sampled" window, ignoring the conditional information of the "context" window actually.
To address this problem, we utilize the "context" window to provide supplemental conditional information, teaching the model to impute the "sampled" window while maintaining the contextual consistency on adjacent windows.
Although the "context" window is accessible during training, it is not available during inference.
Therefore, we incorporate a mixup mechanism~\cite{zhang2018mixup,shen2023timediff} to combine the conditional information in the "sampled" window with that in the "context" window as follows:
\begin{equation}
\label{x_mix}
    \mathbf{x}_{mix}^{co}=\mathbf{m}^k\odot x_{1,t-L:t}^{co}+(\mathbf{1}-\mathbf{m}^k)\odot \mathcal{F}(x_{2,t:t+L}^{co}),
\end{equation}
where $\mathbf{x}_{mix}^{co}$represents the mixed conditional information, $\mathbf{m}^k$ is a mixing coefficient matrix sampled from the uniform distribution $\mathcal{U}(0,1)$, and $\mathcal{F}$ is a convolution function with $1\times 1$ kernel size.
Notably, we only utilize "context" windows during the training stage, while during the inference stage, we set all elements in $\mathbf{m}^k$ to 1.
Then, $\mathbf{x}_{mix}^{co}$ is passed through a linear layer to obtain the hidden representations $\mathbf{c}$: 
\begin{equation}
    \mathbf{c}=\mathrm{Linear}(\mathbf{x}_{mix}^{co}).
\end{equation}
Subsequently, the conditional information representation $\mathbf{c}$, which incorporates information from adjacent windows, is used in the encoder of the denoising network to capture dependencies between observed and missing values.

\subsubsection{Denoising Network}
The denoising network accepts the noised "sampled" window and the conditional information representation $\mathbf{c}$ to predict the noise at missing positions, and then generate the imputed values.

\noindent{\textbf{Embedding:}}
First, the noised "sampled" window $\mathbf{x}_k^{ta}$ is embed on a linear layer, plus with the diffusion step representation $\mathbf{p}^k$ to acquire the hidden representation $h$:
\begin{equation}
    h=\mathrm{Linear}(\mathbf{x}_k^{ta}) +\mathbf{p}^k,
\end{equation}
\begin{equation}
\label{p_k}
    \mathbf{p}^k=\mathrm{FeedForward}(k_{embedding}(k)),
\end{equation}
\begin{equation}
\label{k_embedding}
    \begin{aligned}
    k_{embedding}(k)=[&sin(10^{\frac{0\times 4}{w-1}}k),\cdots,sin(10^{\frac{w\times 4}{w-1}}k),\\
    &cos(10^{\frac{0\times 4}{w-1}}k),\cdots,cos(10^{\frac{w\times 4}{w-1}}k)],     
    \end{aligned}
\end{equation}
where $\mathrm{FeedForward}(\cdot)$ is a two fully-conncted layers with the $SiLU$ activation function, $k_{embedding}$ is $d$-dimension vectors, $w=\frac{d}{2}$.
Then, the $h$ is sent to the encoder to learn the temporal and variable dependencies.

\noindent{\textbf{Encoder:}}
First, a linear layer is utilized to fuse the input, and then, we utilize the single-layer vanilla  transformer block~\cite{vaswani2017attention} to capture the temporal dependency. 
Inspired by the recent work iTransformer~\cite{liu2023itransformer}, we introduce another single-layer invert transformer block to capture the variable dependency.
The encoder is stacked by multiple encoder-layers.
For a single encoder-layer, the dependency extraction process is defined as follows:
\begin{equation}
\begin{aligned}
    \mathbf{H}&=\mathrm{Trm}(\mathrm{Linear}(h)+\mathrm{TE}(h)),\\
    \mathbf{H}_{inv}&=\mathrm{iTrm}(\mathrm{Transpose}(\mathbf{H})+\mathrm{FE}(\mathbf{H}))),
\end{aligned}
\end{equation}
where $\mathbf{H}$ is the output of transformer, $\mathbf{H}_{inv}$ is the output of inverted transformer, $\mathrm{Trm}$ represents the transformer block, $\mathrm{iTrm}$ represents the itransformer block, $\mathrm{TE}(\cdot)$ represents the temporal position embedding, $\mathrm{FE}(\cdot)$ represents the feature position embedding.

\noindent{\textbf{Decoder:}}
The decoder is to merge multiple output from the encoder to acquire $\mathbf{H}_{out}$, and then generate the predicted noise $\hat{\epsilon}$ through a feed-forward network implemented by two fully-connected layers with $ReLU$ activation function:
\begin{equation}
\label{model_output}
\begin{aligned}
    \mathbf{H}_{out}&=\mathrm{LN}(\mathrm{Concat}(\mathbf{H}^{1}_{inv},\cdots,\mathbf{H}^{L}_{inv})),\\
    \epsilon_\theta(\mathbf{x}^{ta}_0,k|\mathbf{c})&=\mathrm{FeedForward}(\mathbf{H}_{out}),
\end{aligned}
\end{equation}
\begin{equation}
\label{hat_epsilon}
    \hat{\epsilon}^{ta}_{k-1}=\frac{1}{\sqrt{\alpha_k}}\mathbf{x}^{ta}_k-\frac{1-\alpha_k}{\sqrt{1-\Bar{\alpha}_k}\sqrt{\alpha}_k}\epsilon_\theta(\mathbf{x}^{ta}_k,k|\mathbf{c})+\sigma_k\epsilon,
\end{equation}
where $\mathbf{H}^{l}_{inv}$ represents the output of $l$-th layer, $L$ is the number of encoder layers.

\subsection{Training}
During the training process, for each predicted noise $\hat{\epsilon}$, we calculate the denoising loss $\mathcal{L}_\epsilon$ as follows:
\begin{equation}
\label{L_epsilon}
\begin{aligned}
     \mathcal{L}_\epsilon&=\mathbb{E}_{\mathbf{x}_0^{ta},\epsilon\sim\mathcal{N}(\mathbf{0},\mathbf{I}),k}\mathcal{L}_\epsilon(k),\\
     \mathcal{L}_\epsilon(k)&=\Vert (1-\mathbf{M})\odot(\epsilon-\epsilon_\theta(\mathbf{x}_k^{ta},k|\mathbf{c}))\Vert_2^2.
\end{aligned}
\end{equation}
Then, the total loss $\mathcal{L}$ is obtained by adding the denoising loss $\mathcal{L}_{\epsilon}$ and the intra contrastive loss $\mathcal{L}_{CL}$ with weighted coefficient $\lambda$:
\begin{equation}
\label{L_theta}
\mathcal{L}=\mathcal{L}_{\epsilon}+\lambda\mathcal{L}_{CL}.
\end{equation}
The complete training procedure is shown in Algorithm~\ref{training}.

\begin{algorithm}
\SetAlgoLined
\KwIn{Incomplete time series $\mathbf{X}$, mask matrix $\mathbf{M}$, the number of diffusion steps $K$, the noise levels $\Bar{\alpha}_k$.}
\KwOut{The optimized denoising network $\epsilon_\theta$}
\Repeat{\rm converged}{
    sample the "sampled" window $x_{t-L:t}$ and its "context" window $x_{t:t+L}$\;
    generate two views by complementary mask based on random mask matrix $m$: $m\odot x_{t-L:t}$ and $(1-m)\odot x_{t-L:t}$ \;
    sample $k \sim $ Uniform({1,$\cdots$,$K$}),  $\epsilon\sim\mathcal{N}(\mathbf{0},\mathbf{I})$ \;  
    obtain the noised $\mathbf{x}^{ta}_k$ using (\ref{noising_target})\;
    obtain the diffusion step embedding using (\ref{p_k}) and (\ref{k_embedding})\;
    obtain the mixed conditional information $\mathbf{x}_{mix}^{co}$ by (\ref{x_mix})\;
    use the $\mathbf{x}_{mix}^{co}$ to acquire conditional embedding $\mathbf{c}$\;
    use the denoising network to predict the noise $\hat{\epsilon}$ by (\ref{model_output}) and (\ref{hat_epsilon})\;
    obtain the embedding $z_1,z_2$ of the contrastive pairs after the encoder of denoising network\;
    calculate the intra contrastive loss $\mathcal{L}_{CL}$ using (\ref{L_intra})\;
    calculate the denoising loss $\mathcal{L}_\epsilon$ using (\ref{L_epsilon})\;
    calculate the total loss $\mathcal{L}$ using (\ref{L_theta})\;
    Update the gradient $\nabla_\theta\mathcal{L}$\;
}
 \caption{Training process of MTSCI.}
 \label{training}
\end{algorithm}

\subsection{Inference}
During the inference process, all elements in the mixing coefficient matrix is set to 1.
By repeatedly running the denoising step till k equals 1, we obtain the $\hat{\epsilon}$ as the final predicted noise.
The inference procedure is shown in Algorithm~\ref{inference}.

\begin{algorithm}
\SetAlgoLined
\KwIn{A sample of incomplete time series $\mathbf{X}$, mask matrix $\mathbf{M}$, the number of diffusion step $K$, the optimized denoising network $\epsilon_\theta$.}
\KwOut{The imputed values $\hat{\mathbf{x}}^{ta}_0$ of the $\mathbf{X}$}
sample the noised $\mathbf{x}^{ta}_k\sim\mathcal{N}(\mathbf{0},\mathbf{I})$\;
\For{$k=K,\cdots,1$}{
$\epsilon\sim\mathcal{N}(\mathbf{0},\mathbf{I})$, if $k>1$, else $\epsilon=0$\;
obtain the diffusion step embedding using (\ref{p_k}) and (\ref{k_embedding})\;
Set all the element of mixing coefficient matrix as 1 and then obtain $\mathbf{x}_{mix}^{co}$ by (\ref{x_mix})\;
use the $\mathbf{x}_{mix}^{co}$ to acquire conditional embedding $\mathbf{c}$\;
predict the $\hat{\mathbf{x}}_{k-1}$ by (\ref{model_output}) and (\ref{hat_epsilon})\;
$\mathbf{x}^{ta}_{k-1}=\hat{\mathbf{x}}^{ta}_{k-1}$\;
}
\Return $\hat{\mathbf{x}}^{ta}_0$\;
 \caption{Inference process of MTSCI.}
 \label{inference}
\end{algorithm}
\section{Experiments}
In this section, we first introduce the experimental setups, including the datasets, baselines, evaluation metrics and implementation details.
Then, we evaluate our model, MTSCI, with extensive experiments to answer the following research questions:

\begin{itemize}[leftmargin=*]
    \item \textbf{RQ1:} How does MTSCI perform against other baselines in the multivariate time series imputation task?
    \item \textbf{RQ2:} How does the imputation consistency of MTSCI contribute to its imputation performance?
    \item \textbf{RQ3:} How does the imputation performance for MTSCI perform about different missing scenarios, including different missing ratios and diverse mask patterns?
    \item \textbf{RQ4:} What is the impact of weighted coefficient $\lambda$ and major hyperparameters of MTSCI on the imputation performance?
\end{itemize}

\subsection{Experimental Setup}\
\label{experimental_setup}
\noindent{\textbf{Datasets.}}
We evaluate our proprosed model on three commonly used public datasets: an electricity dataset \textit{ETT}~\cite{du2023saits}, a climate dataset \textit{Weather}~\cite{wu2021autoformer}, and a traffic speed dataset \textit{METR-LA}~\cite{li2018diffusion}. 
The statistical details of these datasets are listed in Table~\ref{tab:dataset}.

\begin{table}[htbp]
\centering
\caption{Statistical details of datasets.}
\label{tab:dataset}
\resizebox{0.8\columnwidth}{!}{%
\begin{tabular}{@{}c|c|c|c@{}}
\toprule
Datasets              & ETT & Weather & METR-LA  \\ \midrule
Time span              &  69,680     &   52,696      &  34,272            \\
Interval              &  15 min     &   10 min      &  5 min           \\
Features              &  7     &   21      &   207           \\
Sequence length       & 24      &  24       &   24  \\
Original missing rate &  0\%     &   0.017\%      &   8.6\%          \\ \bottomrule
\end{tabular}%
}
\end{table}

\noindent{\textbf{Baselines.}}
We select 13 baselines to evaluate the performance of our proposed method on multivariate time series imputation task.
These baselines include statistical methods (Mean, KNN), typical machine learning methods (MICE, TRMF), deterministic imputation methods (BRITS, mTAN, SAITS, TimesNet, Non-stationary Transformer) and deep generative imputation models (GP-VAE, rGAIN, CSBI, CSDI).
We briefly introduce the baseline methods as follows:
(1)\textbf{Mean}: directly use the average value of observed values to impute. (2)\textbf{KNN}: use the average value of similar samples to missing sample, as implemented by \textit{fancyimpute}.
(3)\textbf{MICE}~\cite{white2011mice}: multiple imputation method by chained equations.
(4)\textbf{TRMF}~\cite{yu2016trmf}: a temporal regularized matrix factorization method. (5)\textbf{GP-VAE}~\cite{fortuin2020gpvae}: combine VAE~\cite{kingma2013vae} with Guassion process for time series probabilistic imputation. (6)\textbf{rGAIN}~\cite{yoon2018gain}: a GAN-based method with a bidirectional recurrent encoder-decoder. (7)\textbf{BRITS}~\cite{cao2018brits}: use bidirectional RNN for multivariate time series imputation. (8)\textbf{mTAN}~\cite{shukla2020mtan}: use multi-time attention network to impute missing values, which is a transformer-based method. (9)\textbf{SAITS}~\cite{du2023saits}: use joint-optimization training strategy (masked imputation task and observed reconstruction task) to impute missing values, which is also a tranformer-based method. (10)\textbf{Non-stationary Transformer}~\cite{liu2022nonstationary}: a tranformer-based method to attenuate time series non-staionary. (11)TimesNet~\cite{wu2023timesnet}: transform the 1D time series into 2D space and capture the temporal 2D-variations dependencies. Both Non-stationary Transformer and TimesNet are the state-of-the-art multivariate time series imputation methods implemented in TSlib.
(12)\textbf{CSBI}~\cite{chen2023csbi}: use the Schr\"odinger bridge algorithm to probabilistic time series imputation. (13)\textbf{CSDI}~\cite{tashiro2021csdi}: a score-based conditional diffusion model for probabilistic time series imputation method.

\noindent{\textbf{Evaluation metrics.}}
Three commonly used metrics in multivariate time series imputation task are used to evaluate the performance of all methods, including the Mean Absolute Error(MAE), Root Mean Squared Error(RMSE), Mean Absolute Percentage Error(MAPE). 

\noindent{\textbf{Implementation.}}
We divide the training/validation/testing set following the settings of previous works~\cite{du2023saits,wu2021autoformer,liu2023pristi}. 
For ETT, we select the first four-month data(2016/07-2016/10) as the testing set, the following four-month data(2016/11-2017/02) as the validation set, and the left sixteen months(2017/03-2018/06) as the training set. 
For Weather and METR-LA, we split the training/validation/testing set by 70\%/10\%/20\%. 
We divide the samples by a window size of 24 steps without overlapping.
We consider two different missing patterns to artificially simulate the missing values for evaluation: (1)\textbf{Point missing:} we randomly mask 20\% data points in the datasets. 
(2)\textbf{Block missing:} based on randomly masking 5\% of the observed data, mask observations ranging from [L/2, 2L] (L is the window size) with 0.15\% probability.
For training strategies, we utilize point and block strategies for self-supervised learning. 
Specifically, for point strategy, we randomly choose $r$ ($r\in [0\%,100\%]$) of observed values as imputation targets. 
For block strategy, we randomly choose a sequence with a length in the range [L/2,L] with a probability $r$ ($r\in [0\%,15\%]$) as imputation targets.


\begin{table*}[!t]
\caption{Overall performance on three datasets with point missing pattern. Performance averaged over 5 runs.}
\label{tab:pointmissing}
\resizebox{\textwidth}{!}{%
\begin{tabular}{@{}cccccccccc@{}}
\toprule
\multirow{2}{*}{Methods} & \multicolumn{3}{c}{ETT} & \multicolumn{3}{c}{Weather} & \multicolumn{3}{c}{METR-LA} \\ \cmidrule(l){2-10} 
                         & MAE    & RMSE    & MAPE   & MAE      & RMSE      & MAPE     & MAE     & RMSE    & MAPE    \\ \midrule
Mean                     &  $3.676\pm 0.000$         &  $6.557\pm 0.000$         & $46.451\pm 0.000 $         &$ 38.331\pm 0.000 $         & $89.201\pm 0.000 $         &  $22.726\pm 0.000$          &  $7.136\pm 0.000$         & $11.272\pm 0.000$          & $12.458\pm 0.000 $     \\
KNN                   &  $3.561\pm0.000$   &   $6.708\pm 0.000$      &    $44.994\pm 0.000 $    &   $5.002\pm 0.000  $   &  $25.454\pm 0.000$  & $2.965\pm 0.000 $       &  $8.365\pm  0.000 $       &   $14.473\pm  0.000 $     &   $14.592\pm  0.000 $               \\
MICE                     &  $2.324\pm0.000 $      & $5.586\pm  0.000$       &  $29.371\pm 0.000$      &   $12.834\pm  0.000$      &    $46.177\pm0.000 $        &   $7.609\pm0.000$        &   $3.301\pm 0.000 $     &   $ 5.125\pm 0.000 $    &    $5.763\pm 0.000$      \\
TRMF                     &  $ 2.023 \pm 0.000$     &$ 3.213 \pm 0.000$        &  $25.646 \pm 0.004$      & $30.106 \pm 0.176 $        & $75.557 \pm 0.412$          &  $17.871 \pm 0.105$        &   $5.908 \pm 0.002$      &  $8.416 \pm 0.003 $      & $10.228 \pm 0.003 $       \\
GP-VAE                   & $ 1.446 \pm 0.084 $     &  $2.474 \pm 0.113$       &  $17.376 \pm 0.982 $     &  $9.216 \pm 0.602 $       &  $27.112 \pm 1.032$         &  $5.470 \pm 0.357 $       & $ 3.305 \pm 0.040 $      &   $5.312 \pm 0.020  $    &   $5.788 \pm 0.071 $     \\
rGAIN                    &$ 0.804 \pm 0.032$       &  $1.743 \pm 0.071$       & $10.167 \pm 0.399 $      &  $7.597\pm0.129 $       &  $25.139\pm 0.230 $        &   $4.505\pm 0.077$       &  $ 3.115\pm 0.010 $     &  $4.810 \pm 0.014 $      &  $ 5.439 \pm 0.017 $     \\
BRITS                    &   $0.634\pm 0.041 $    &  $1.491 \pm0.115$       &  $8.006 \pm 0.516$      &$ 8.487\pm0.317$         &   $25.844\pm 0.842$        &  $5.032\pm  0.188 $       &  $2.981\pm 0.009$       &  $4.672\pm0.011$       &   $5.205 \pm0.015 $     \\
mTAN                     & $0.525 \pm 0.114  $     &  $0.984 \pm 0.185$       &  $6.654 \pm 1.444$      &    $5.872 \pm 0.215$      &    $21.498 \pm 0.271 $      &    $3.483 \pm 0.128 $     & $3.611 \pm 0.091 $       & $ 5.681 \pm 0.203  $     &  $6.306 \pm 0.160 $      \\
SAITS                    & $0.460\pm 0.049$       & $0.832\pm 0.121 $       &   $5.812 \pm 0.615 $    &  $4.113\pm 0.108$        &  $19.207\pm 0.171$         &    $2.438 \pm 0.064 $      &  $2.543 \pm 0.012$       &   $ 4.381\pm 0.008$     &   $4.441 \pm 0.021$      \\
TimesNet                 &  $0.317 \pm0.002$      &   $0.511\pm0.005 $     &  $4.007\pm0.027$      &   $4.114 \pm 0.121 $      &  $ 20.916 \pm 0.242$        &  $2.439 \pm 0.071$        &  $2.474\pm 0.004$       &  $4.346\pm 0.006$       &   $4.321 \pm 0.006 $     \\
Stationary               &  $0.297 \pm 0.003 $     &  $0.481 \pm 0.006 $      &  $3.754 \pm 0.035$      &  $4.415\pm 0.748$        &   $21.733\pm 2.763 $       &  $2.618\pm 0.443$        & $2.502 \pm 0.006   $     &   $4.402 \pm 0.009$      &   $4.369  \pm 0.010$      \\
CSBI                     &   $0.270 \pm 0.014$     &   $0.467 \pm 0.035$      &   $3.262 \pm 0.164$     &   $3.506 \pm 0.277$       &    $23.108 \pm 2.354$       &  $1.950 \pm 0.154$        &  $2.683 \pm 0.010 $      & $4.702 \pm 0.052$        &   $4.751 \pm 0.017$      \\
CSDI                     &   $0.225\pm 0.000 $    &   $0.383\pm 0.001$      &  $2.838\pm 0.004$      &   $2.084\pm 0.007 $      & $17.003\pm 0.135$          &   $1.236\pm 0.004$       &   $1.733\pm 0.000$      &  $3.248\pm 0.001$       &   $ 3.026\pm 0.000$     \\ \midrule
\textbf{MTSCI(ours)}                     & \bm{$0.214\pm 0.001$}       & \bm{$0.358\pm 0.003$}       &\bm{$2.711\pm 0.020$}       & \bm{$1.955\pm 0.011$}        & $\bm{16.162\pm 0.186}$        &\bm{$1.160\pm 0.006$}         & \bm{$1.655\pm 0.011$}       & \bm{$3.076\pm 0.026$}       &\bm{$2.889\pm 0.019$}        \\ 
\bottomrule
\end{tabular}%
}
\end{table*}

\begin{table*}[!t]
\caption{Overall performance on three datasets with block missing pattern. Performance averaged over 5 runs.}
\label{tab:blockmissing}
\resizebox{\textwidth}{!}{%
\begin{tabular}{@{}cccccccccc@{}}
\toprule
\multirow{2}{*}{Methods} & \multicolumn{3}{c}{ETT} & \multicolumn{3}{c}{Weather} & \multicolumn{3}{c}{METR-LA} \\ \cmidrule(l){2-10} 
                         & MAE    & RMSE    & MAPE   & MAE      & RMSE      & MAPE     & MAE     & RMSE    & MAPE    \\ \midrule
Mean                     & $3.407\pm 0.000$         & $6.248\pm 0.000 $        & $46.235\pm 0.000$          & $41.383\pm 0.000$          &$ 94.105\pm 0.000$          & $23.057\pm 0.000$          &  $7.120\pm 0.000$         & $11.229\pm 0.000 $         & $12.437\pm 0.000$      \\
KNN                      & $ 2.701\pm0.000 $     &   $6.130\pm 0.000$      & $36.652\pm0.000$       &   $21.245\pm0.000 $       &  $ 71.288\pm0.000$         &   $11.837\pm0.000$       &   $9.012\pm0.000$      &   $14.912\pm0.000$      &    $15.721\pm0.000$     \\
MICE                     & $1.973\pm0.000$       &   $5.212\pm0.000 $     &  $26.783\pm0.000$      &   $10.023\pm0.000$       &  $34.550\pm0.000$         &  $5.584\pm0.000$        &    $8.668\pm0.000$      & $ 11.697\pm0.000$       &   $15.143\pm0.000$        \\
TRMF                     &$ 1.863 \pm 0.012 $      &  $3.156 \pm 0.018$       & $25.329 \pm 0.157$       &  $31.619 \pm 0.136$        &  $77.851 \pm 0.350$         &  $17.620 \pm 0.076$        & $5.781 \pm 0.001$        &  $ 8.299 \pm 0.000 $     &  $10.009 \pm 0.001$       \\
GP-VAE                   &$ 2.365 \pm 0.100$     &  $4.344 \pm 0.100 $      &  $29.726 \pm 1.124 $     &  $14.579 \pm 0.034 $       &    $38.357 \pm 0.344$       &   $8.493 \pm 0.084 $      & $ 3.393 \pm 0.008 $      &  $5.478 \pm 0.013 $      & $5.942 \pm 0.013 $       \\
rGAIN                    & $1.964\pm0.099$       &  $4.837 \pm 0.159$       &   $26.652 \pm 1.347$      &  $ 11.359\pm 0.118 $      &   $ 33.929\pm 0.301 $      &   $6.329\pm 0.065$       &   $3.604\pm0.014$      &    $5.592\pm0.020 $     &  $6.297 \pm0.024$       \\
BRITS                    &  $1.676 \pm 0.055 $     &   $4.790 \pm 0.124$      &   $22.744 \pm 0.745$     &  $13.239\pm 0.949 $       &   $37.372\pm2.234 $       &  $7.376\pm 0.529 $       &   $3.352\pm0.014$      &  $5.372\pm 0.025 $       &  $5.856\pm 0.025$       \\
mTAN                     & $ 1.284 \pm 0.070$      &    $3.558 \pm 0.035$     & $17.478 \pm 0.949$       &  $11.411 \pm 0.528$        &  $38.816 \pm 1.621$         & $6.351 \pm 0.294$         & $3.809 \pm 0.060$        &  $6.605 \pm 0.141$       &$ 6.656 \pm 0.106$        \\
SAITS                    &   $1.475 \pm 0.060$     &$ 4.232 \pm 0.117  $      &   $20.011 \pm 0.807$     &   $7.286\pm 0.169 $      &   $28.784\pm 0.157 $        &   $4.059 \pm 0.094$        &   $2.928 \pm  0.012  $    &   $5.163\pm 0.018 $     &   $5.115\pm 0.021$      \\
TimesNet                 &  $1.553  \pm 0.005$      &  $ 3.940 \pm 0.004$      & $21.072\pm 0.065$       &    $21.915\pm 0.311$      &  $71.433\pm 0.619$         &   $12.210\pm 0.173 $     &  $5.140\pm 0.007 $      &    $8.629 \pm 0.011 $    &    $8.980\pm 0.012 $     \\
Stationary               & $1.525 \pm 0.005$       &   $3.926 \pm 0.004$      &    $20.692 \pm 0.064$    &  $ 21.808\pm 0.083 $      & $71.322\pm 0.207$          &  $12.151\pm  0.046$        &  $ 5.139\pm 0.004$      &  $ 8.626\pm 0.010 $      &   $8.978\pm 0.007 $     \\
CSBI                     &  $1.261 \pm 0.064 $     &  $4.365 \pm 0.230 $      &  $14.604\pm 0.703$      &  $18.653 \pm 1.082$        &   $63.743 \pm 4.381 $       &  $13.440 \pm 0.780$        &   $5.574 \pm 0.248 $     &  $9.665 \pm 0.240 $      &  $9.904 \pm 0.440$       \\
CSDI                     &  $0.967\pm 0.004$      &   $3.580\pm 0.028$      &   $13.120\pm 0.058$     & $4.648\pm 0.021$         &   $26.598\pm 0.053$        &   $2.590\pm 0.011$       &   $2.582\pm 0.002$      &   $5.534\pm 0.006$      &  $4.511\pm 0.003 $      \\ \midrule
\textbf{MTSCI(ours)}                     &   \bm{$0.642\pm 0.032$}     & \bm{$2.706\pm 0.101$}        & \bm{$8.350\pm 0.418$}       &\bm{$ 3.092\pm 0.035 $}       &   \bm{$ 21.267\pm 0.159$}      & \bm{$2.407\pm 0.063$ }       & \bm{$1.982\pm 0.016$}       & \bm{$3.914\pm 0.027$}       & \bm{$3.462\pm 0.028$}       \\
\bottomrule
\end{tabular}%
}
\end{table*}

\vspace{-2mm}
\subsection{Overall Performance(RQ1)}
The overall performance is shown in Table~\ref{tab:pointmissing} and Table~\ref{tab:blockmissing}. 
We make the following observations: 
(1) MTSCI achieves the state-of-the-art performance across multiple datasets, both in point missing and block missing patterns, with an average improvement of 17.88\% in MAE, 15.09\% in RMSE and 13.64\% in MAPE, respectively.
Notably, in the case of block missing pattern with continuous missing scenarios, our method demonstrates greater superiority, highlighting the effectiveness of the imputation consistency strategy adopted in MTSCI.
(2) The statistical methods and classical machine learning methods perform poor on all datasets due to the strong non-linearity of incomplete multivariate time series.
These methods impute missing values based on assumptions such as stability or linear dynamics, which fail to capture the complex temporal correlations in real-world datasets. 
(3) Compared to deterministic imputation models, MTSCI achieves performance improvements of 42.07\% in MAE, 24.15\% in RMSE and 39.76\% in MAPE respectively on average.
The difference between these methods and our model is that we employ the conditional diffusion process to model the incomplete time series imputation task, which refines the imputed values through multiple steps instead of non-autoregressive single step imputation as deterministic methods do.
(4) Compared with deep generative imputation methods, MTSCI consistently outperforms on several datasets.
This indicates that our imputation consistency strategy effectively enhances imputation performance.
Previous generative models, including the conditional diffusion models like CSBI and CSDI, still exhibit significant errors because they rely solely on self-supervised masking strategy to generate imputation targets and directly guide the denoising network through the inductive bias at the imputation targets.
In contrast, MTSCI utilizes a complementary mask strategy to generate dual views for intra contrastive loss and a mixup mechanism to combine conditional information from adjacent windows, facilitating more accurate and consistent imputation performance.

\vspace{-2mm}
\subsection{Ablation Study(RQ2)}
We conduct an ablation study to evaluate the effectiveness of the complementary mask strategy to generate intra contrastive loss and the inter-consistency condition network with mixup mechanism for utilizing conditional information from adjacent windows.
We compare three variants of MTSCI with and without these components:
(1) We remove the complementary mask strategy on the forward noising process along with the intra-consistency loss. This variant is denoted as \textit{w/o intra}.
(2) We remove the mixup mechanism in inter-consistency condition network, which is denoted as \textit{w/o inter}.
(3) We use only the conditional information of observed values from single window and the denoising network of MTSCI, without the complementary mask strategy and mixup mechanism. This variant is denoted as \textit{w/o cons}.

\begin{figure}[!t]
    \centering
    \includegraphics[width=\linewidth]{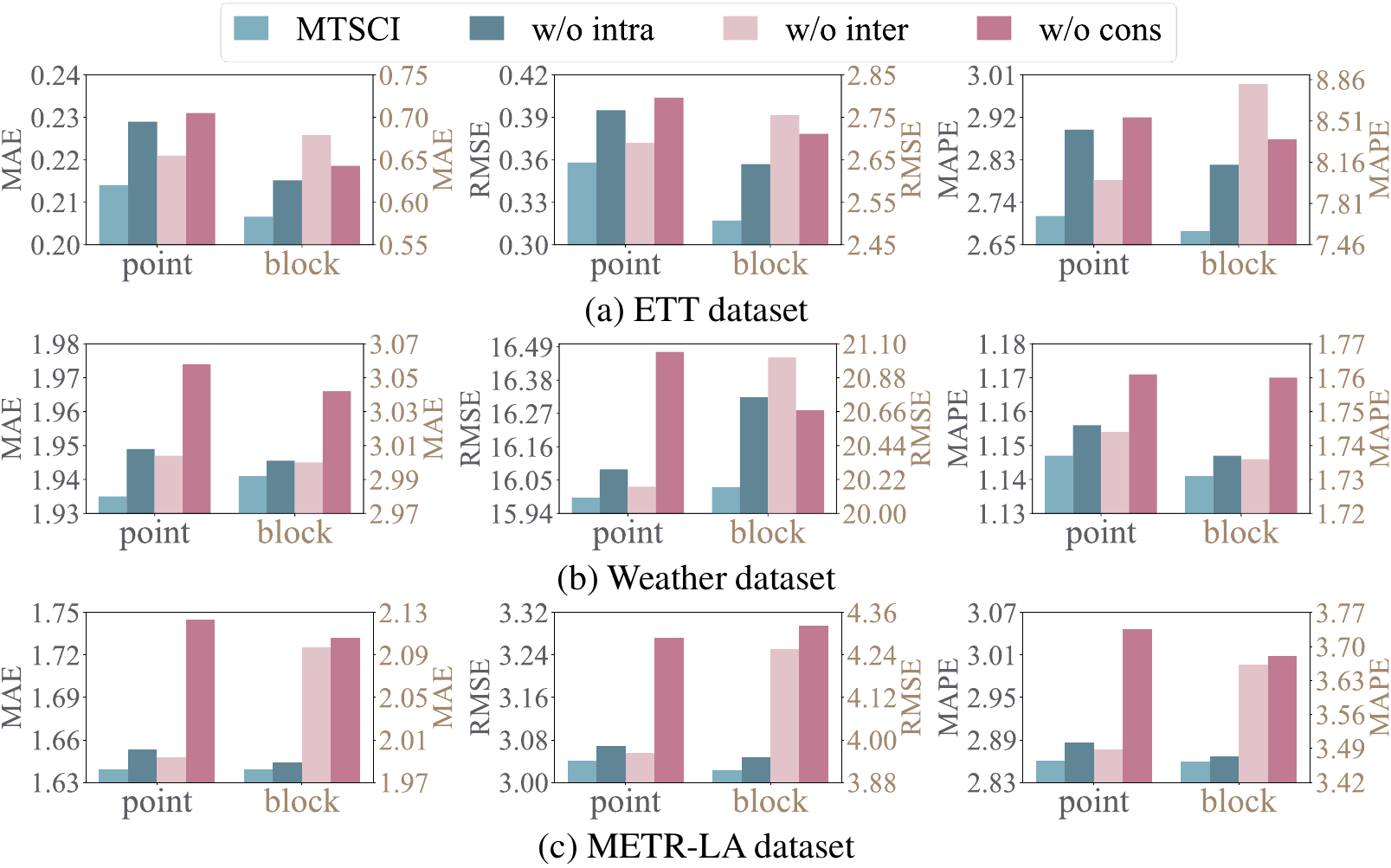}
    \caption{Ablation study of MTSCI. 
    }
    \label{fig:ablation_study}
    \vspace{-2mm}
\end{figure}

Figure~\ref{fig:ablation_study} shows the performance comparison of these three variants and MTSCI.
First, we observe that without the complementary mask strategy, the performance of this variant deteriorates.
This indicates that generating contrastive views to facilitate the mutual reconstruction of observed and imputed values, improves the imputation performance.
Second, using adjacent windows to provide supplemental conditional information also enhances the imputation performance.
This suggests that the adjacent windows can bring contextual consistency constraints to the missing values of the current window, alleviating the estimation error between the imputation results and the ground-truth.
Finally, the collaboration of these modules jointly improves the imputation performance, further confirming the necessity of utilizing both simultaneously. 

We also conduct experiments to compare the performance using the same denoising network architecture but with different objectives: predict noise $\epsilon$ or predict missing values $\mathbf{x}_0^{ta}$.
As shown in Table~\ref{tab:denoising_strategy}, performance is better when the denoising network's objective is to predict noise.
This is likely because noise follows a Gaussian distribution, which aids the complementary mask views in predicting noise and reconstructing each other, thereby maintaining the intra-consistency.
However, the distributions of observed values and missing values may not be the same.

\begin{table}[!t]
\centering
\caption{RMSEs of two different denoising objectives: Predicting 
 noise $\epsilon_{\theta}$ vs Predicting target $x_{\theta}$.}
\label{tab:denoising_strategy}
\resizebox{\columnwidth}{!}{%
\begin{tabular}{@{}ccccccc@{}}
\toprule
\multirow{2}{*}{Methods} & \multicolumn{2}{c}{ETT} & \multicolumn{2}{c}{Weather} & \multicolumn{2}{c}{METR-LA} \\ \cmidrule(l){2-7} 
          & Point & Block & Point & Block & Point & Block \\ \midrule
Predicting $\epsilon_{\theta}$      &  \textbf{0.358}     &  \textbf{2.706}     &  \textbf{16.162}     &  \textbf{21.267}     &  \textbf{3.076}     & \textbf{3.914}      \\
Predicting $x_{\theta}$ &  0.775     &  2.836     &  22.739     &  30.847     &  4.111     &  5.523     \\
\bottomrule
\end{tabular}%
}
\vspace{-2mm}
\end{table}

\vspace{-2mm}
\subsection{Case Study(RQ2)}
\begin{figure*}[!t]
    \centering
    \includegraphics[width=0.9\linewidth]{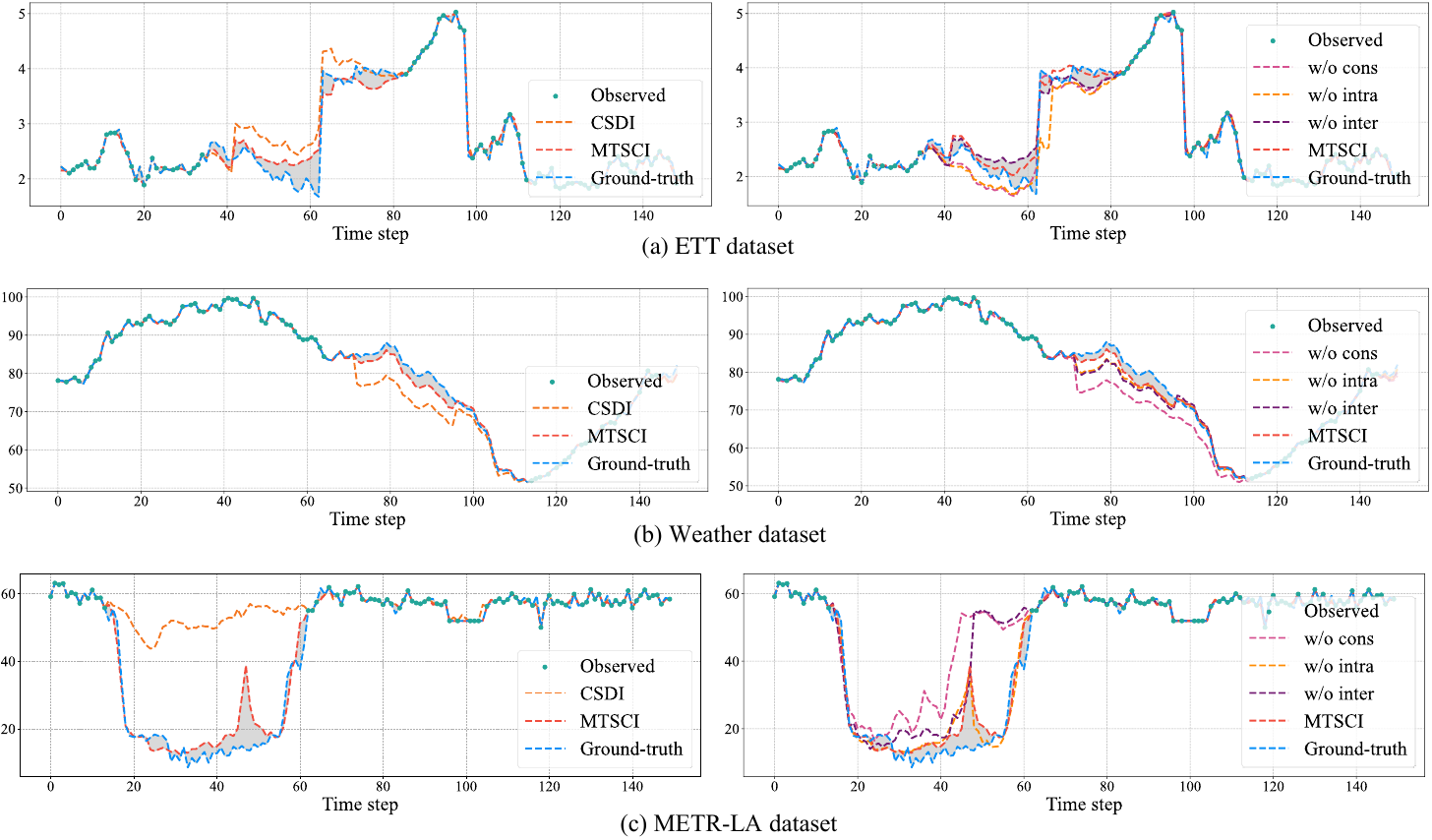}
    \caption{The visualization of the imputation results of snapshots on three datasets: MTSCI vs CSDI and MTSCI vs three variants.}
    \label{fig:case_study}
    \vspace{-2mm}
\end{figure*}

In order to intuitively understand how MTSCI imputes the incomplete time series, we visualize the imputation results of MTSCI and the sub-optimal method CSDI.
Specifically, we randomly select three snapshots of incomplete time series with block missing pattern from three datasets.
As shown in Figure~\ref{fig:case_study}, MTSCI demonstrates better imputation performance compared to the CSDI, which is attributed to intra contrastive consistency and inter-consistency condition network.
In addition, compared to three variants of MTSCI, our model exhibits consistent trend between imputed and observed values, alleviating the imputation error.
This improvement is due to MTSCI's consideration of intra-consistency within a single window that observed and imputed values can reconstruct each other, as well as the inter-consistency between adjacent windows that can provide supplemental condition information to guide imputation.
Additionally, we use the CRPS metric~\cite{matheson1976crps,tashiro2021csdi} to measure the imputation consistency between the imputed results and the observed values at the whole dataset. 
As shown in Table~\ref{tab:crps}, our method outperforms the sub-optimal method CSDI.
\vspace{-2mm}
\label{case_study}

\begin{table}[!t]
\centering
\caption{The imputation consistency, evaluated by CRPS~\cite{matheson1976crps}, are presented for both MTSCI and CSDI. Lower values indicate better consistency performance.}
\label{tab:crps}
\resizebox{\columnwidth}{!}{%
\begin{tabular}{@{}ccccccc@{}}
\toprule
\multirow{2}{*}{Methods} & \multicolumn{2}{c}{ETT}           & \multicolumn{2}{c}{Weather}       & \multicolumn{2}{c}{METR-LA}       \\ \cmidrule(l){2-7} 
     & Point  & Block  & Point  & Block  & Point  & Block  \\ \midrule
MTSCI                    & \textbf{0.0206} & \textbf{0.0626} & \textbf{0.0085} & \textbf{0.0137} & \textbf{0.0220} & \textbf{0.0265} \\
CSDI & 0.0220 & 0.0652 & 0.0095 & 0.0164 & 0.0233 & 0.0383 \\ \bottomrule
\end{tabular}%
}
\vspace{-2mm}
\end{table}

\subsection{Sensitivity Analysis(RQ3)}
To evaluate the generalization ability of MTSCI, we carry out an assessment of performance w.r.t. different missing ratios on Weather dataset (due to space limitations).
To comprehensively account for the sufficiency and sparsity of missing data, we use missing ratios of {10\%,30\%,50\%,70\%} in testing set.
We compare MTSCI with two other conditional diffusion models: CSBI and CSDI.
The results are shown in Table~\ref{tab:sensitivity}.
MTSCI outperforms the baselines across different missing ratios both on two different mask patterns.
However, we observe that the imputation performance does not consistently decrease with increasing missing ratios, indicating the presence of distribution shift in incomplete time series.
Notably, our consistency strategy helps alleviate this issue to some extent, maintaining good generalization ability.
To further verify the generalization of our model, we also evaluate it under different missing patterns for the training and testing sets. 
Specifically, we use two settings: \textbf{Point-\textgreater{}Block} (Point missing pattern in training set, Block missing pattern in testing set.) and \textbf{Block-\textgreater{}Point} (Block missing pattern in training set, Point missing pattern in testing set.)
As shown in Table~\ref{tab:mask_pattern_ood}, our method achieves relatively better performance even when the missing patterns in the training and testing sets differ.
This indicates that our trained model can handle imputation tasks in testing environments with missing patterns that are different from those in the training environment.
\begin{table}[!t]
\centering
\caption{Imputation performance comparison w.r.t different missing ratios on Weather dataset.}
\label{tab:sensitivity}
\resizebox{\columnwidth}{!}{%
\begin{tabular}{@{}cc|ccc|ccc|ccc@{}}
\toprule
\multicolumn{2}{c|}{Methods}                  & \multicolumn{3}{c|}{\textbf{MTSCI}} & \multicolumn{3}{c|}{CSDI} & \multicolumn{3}{c}{CSBI} \\ \midrule
\multicolumn{2}{c|}{Miss Ratio} & MAE   & RMSE   & MAPE  & MAE   & RMSE   & MAPE  & MAE    & RMSE   & MAPE  \\ \midrule
\multicolumn{1}{c|}{\multirow{4}{*}{\rotatebox{90}{Point}}} & 10\% & \textbf{1.995}  & 16.831  & \textbf{1.186}  & 2.123  & \textbf{15.524}  & 1.265  & 4.253  & 30.421 & 2.423  \\
\multicolumn{1}{c|}{}   & 30\%  & \textbf{1.962} & 15.711 & \textbf{1.157} & 2.077 & \textbf{14.952} & 1.225 & 3.875  & 26.488 & 2.185 \\
\multicolumn{1}{c|}{}   & 50\%  & \textbf{2.474} & 18.417 & \textbf{1.464} & 2.569 & \textbf{16.923} & 1.520 & 4.714  & 31.090 & 2.710 \\
\multicolumn{1}{c|}{}   & 70\%  & \textbf{3.546} & \textbf{21.286} & \textbf{2.109} & 4.460 & 24.214 & 2.647 & 5.873  & 35.816 & 3.375 \\ \midrule
\multicolumn{1}{c|}{\multirow{4}{*}{\rotatebox{90}{Block}}} & 10\% & \textbf{3.580}  & \textbf{23.496}  & \textbf{2.041}  & 4.314  & 24.513  & 2.459  & 17.875 & 63.623 & 11.649 \\
\multicolumn{1}{c|}{}   & 30\%  & \textbf{2.714} & \textbf{19.280} & \textbf{1.579} & 3.021 & 20.064 & 1.757 & 10.783 & 47.616 & 6.394 \\
\multicolumn{1}{c|}{}   & 50\%  & \textbf{2.877} & \textbf{20.304} & \textbf{1.694} & 2.951 & 20.458 & 1.738 & 9.377  & 43.415 & 5.518 \\
\multicolumn{1}{c|}{}   & 70\%  & \textbf{3.495} & \textbf{22.287} & \textbf{2.073} & 3.579 & 23.260 & 2.232 & 9.598  & 43.918 & 5.574 \\ \bottomrule
\end{tabular}%
}
\vspace{-2mm}
\end{table}

\begin{table}[!t]
\centering
\caption{Performance comparison of different missing patterns during training and testing.}
\label{tab:mask_pattern_ood}
\resizebox{\columnwidth}{!}{%
\begin{tabular}{@{}ccccccccccc@{}}
\toprule
\multirow{2}{*}{Settings} & \multirow{2}{*}{Methods}  & \multicolumn{3}{c}{ETT} & \multicolumn{3}{c}{Weather} & \multicolumn{3}{c}{METR-LA} \\ \cmidrule(l){3-11} 
                         &                        &MAE   & RMSE        & MAPE      &MAE  & RMSE         & MAPE      &MAE   & RMSE         & MAPE         \\ \midrule
\multirow{3}{*}{\begin{tabular}[c]{@{}c@{}}\textbf{Point}\\$\downarrow$\\ \textbf{Block}\end{tabular}}    & MTSCI  &  0.707 &  \textbf{2.857}       &  9.191    &  \textbf{6.422} &  \textbf{26.739}    &  \textbf{3.716}  &  \textbf{3.478}  &  6.485    &\textbf{ 6.075}           \\
                                                       & CSDI  &  0.936&  3.573   &  12.700    &  14.340&   67.415    &   8.297  &  3.485  &   \textbf{6.457}     &   6.087         \\
                                                       & CSBI  &  \textbf{0.705}&  2.881   &  \textbf{8.385}    &  11.566&   53.313    &   6.968  &  3.928  &   7.962     &   6.967         \\\midrule
\multirow{3}{*}{\begin{tabular}[c]{@{}c@{}}\textbf{Block}\\$\downarrow$\\ \textbf{Point}\end{tabular}}    & MTSCI   &  0.345&  0.679       &  4.360  & \textbf{ 1.899 }  &  \textbf{15.885}     &  \textbf{1.126}   &  \textbf{1.783}  &  \textbf{3.363}     &  \textbf{3.113}    \\
                                                       & CSDI  &  \textbf{0.229}&  \textbf{0.385}   &  \textbf{2.893}  &  2.209   &  17.916      &  1.310  &  1.988   &   3.736      &   3.472  \\ 
                                                       & CSBI  &  0.401&  0.735   &  4.850    &  4.616&   26.578    &   2.567  &  2.790  &   4.766     &   4.941         \\\bottomrule
\end{tabular}%
}
\vspace{-2mm}
\end{table}

\begin{figure}[!t]
    \centering
    \includegraphics[width=\linewidth]{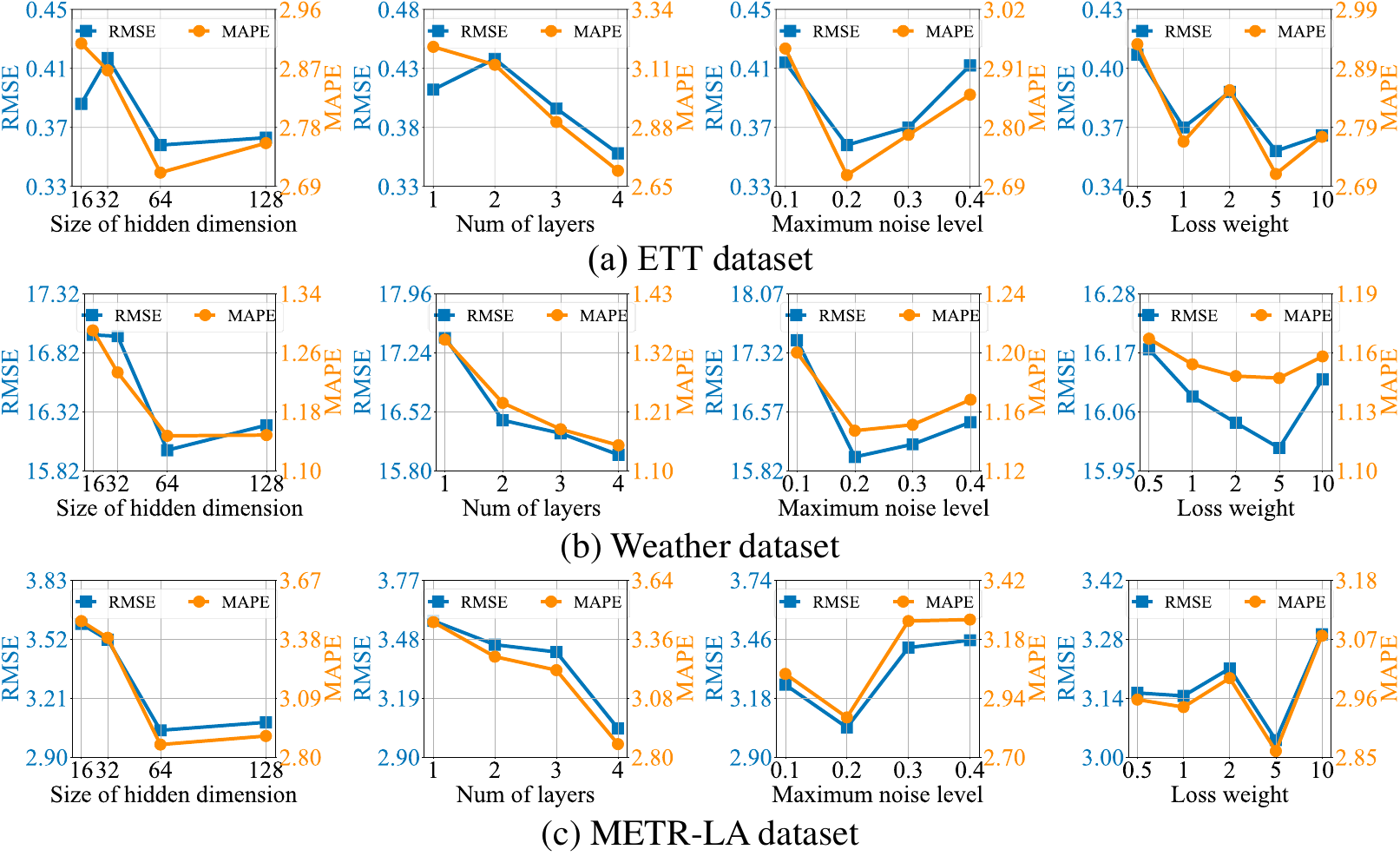}
    \caption{Hyperparameter study on four key parameters of MTSCI: the size of hidden dimension $d_h$, the number of encoder-layer $l$ in denoising network, the maximum noise level $\beta_K$ and the weighted coefficient $\lambda$ of intra-consistency loss. Both RMSE and MAPE are reported.}
    \label{fig:hyperparameter_study}
    \vspace{-3mm}
\end{figure}
\subsection{Hyperparameter Study(RQ4)}
\label{hyperparameter_study}
We conduct a hyperparameter study on key parameters in MTSCI to select the optimal settings across three datasets: the size of hidden dimension $d$, the number of encoder layers $l$ in the denoising network, the maximum noise level $\beta_T$ and the weighted coefficient $\lambda$ of the intra contrastive loss.
The results are shown in Figure~\ref{fig:hyperparameter_study}.
$d$ affects the representation ability of the model, leading to poor performance if it is too large or too small.
While a large $l$ enhances the ability to capture temporal and variable dependencies, thereby improving imputation performance, it also increases the amount of model parameters and computation complexity.
For the level of sampled noise, a moderate value is more conducive to the noise prediction in the denoising network.
To balance the learning objectives of MTSCI during training, we need to adjust the weight coefficient of the contrastive loss on a small scale for better performance.

\vspace{-2mm}
\section{Conclusion}
In this paper, 
we systematically summarize the imputation consistency for improving imputation performance, including the \textit{intra-consistency} and \textit{inter-consistency}.
We propose a conditional diffusion model for multivariate time series consistent imputation, \textbf{MTSCI}.
Specifically, we adopt a complementary mask strategy to introduce intra contrastive loss, ensuring the mutual consistency between the imputed and observed values.
Moreover, we utilize an inter-consistency condition network with a mixup mechanism to incorporate the conditional information from adjacent windows to facilitate imputation.
Extensive experiments demonstrate the effectiveness of our method, achieving state-of-the-art performance on multiple real-world datasets under various experimental settings.
In the future, we will extend our method to apply to more complex missing data scenarios.

\vspace{-2mm}
\begin{acks}
We thank the kind help from Pan Liu (SJTU).
This work was sponsored by National Key Research and Development Program of China under Grant No.2022YFB3904204, National Natural Science Foundation of China under Grant No. 62102246, 62272301, and Provincial Key Research and Development Program of Zhejiang under Grant No. 2021C01034.
\end{acks}


\bibliographystyle{ACM-Reference-Format}
\bibliography{reference}


\end{sloppypar}
\end{document}